\documentclass[conference]{IEEEtran}
\IEEEoverridecommandlockouts
\usepackage{stfloats}
\usepackage{placeins}
\usepackage{cite}
\usepackage{amsmath,amssymb,amsfonts}
\usepackage{algorithmic}
\usepackage{graphicx}
\usepackage{textcomp}
\usepackage{xcolor}
\usepackage{multirow}
\usepackage{amsmath}
\usepackage{caption}
\usepackage{tabularx}
\usepackage{url}
\usepackage[table]{xcolor}
\usepackage{colortbl}
\usepackage[most]{tcolorbox}
\usepackage{enumitem}
\definecolor{systemframe}{HTML}{558B2F}  
\definecolor{systemback}{HTML}{C5E1A5}  
\definecolor{userframe}{HTML}{558B2F}  
\definecolor{userback}{HTML}{C5E1A5}  
\usepackage{adjustbox}
\usepackage{subcaption}
\usepackage[most]{tcolorbox}

\tcbset{
    promptbox/.style={
        colback=gray!5,
        colframe=gray!60,
        boxrule=0.4pt,
        arc=2pt,
        left=6pt,
        right=6pt,
        top=4pt,
        bottom=4pt
    }
}
\def\BibTeX{{\rm B\kern-.05em{\sc i\kern-.025em b}\kern-.08em
    T\kern-.1667em\lower.7ex\hbox{E}\kern-.125emX}}
\begin{document}

\title{FoodCHA: Multi-Modal LLM Agent for Fine-Grained Food Analysis}

\author{
  Woojin Lee,
  Pranav Mekkoth,
  Ye Tian,
  Onat Gungor,
  Tajana Rosing\\
  \textit{Department of Computer Science and Engineering} \\
  \textit{University of California, San Diego (UCSD)} \\
  \{wol027, pmekkoth, yet002, ogungor, tajana\}@ucsd.edu
}

\maketitle 

\begin{abstract}
The widespread adoption of camera-equipped mobile devices and wearables has enabled convenient capture of meal images, making food recognition a key component for real-time dietary monitoring. However, real-world food images present challenges due to high intra-class similarity and the frequent presence of multiple food items within a single image. While deep learning models achieve strong performance in coarse-grained classification, they often struggle to capture fine-grained attributes such as cooking style. Moreover, open-ended generation in modern vision-language models can produce non-canonical labels, limiting their practical deployment. We propose \textbf{FoodCHA}, a multimodal agentic framework that reformulates food recognition as a hierarchical decision-making process. By progressively anchoring predictions, FoodCHA guides subcategory identification using high-level categories and guides cooking style recognition using subcategories, improving semantic consistency and attribute-level discrimination. To ensure practical deployability, \textbf{FoodCHA} utilizes the compact Moondream-2B vision-language model, which provides strong reasoning capability while maintaining lower computational and memory overhead. Experiments on FoodNExTDB show that \textbf{FoodCHA} outperforms Food-Llama-3.2-11B by 13.8\% and 38.2\% in category and subcategory recognition precision, respectively, and achieves a striking 153.2\% improvement in cooking style classification precision. 
\end{abstract}


\begin{IEEEkeywords}
Food recognition, Multimodal learning, Vision-language models, Fine-grained classification
\end{IEEEkeywords}

\section{Introduction}
Camera-equipped mobile devices and wearables have made meal image capture increasingly practical for dietary monitoring and personalized nutrition support~\cite{anikwe2022mobile,turrini2022perspectives}. However, converting real-world meal photos into reliable dietary logs remains difficult. Food images vary substantially in appearance, portion size, preparation style, lighting, and background, while visually similar or mixed dishes may blur semantic boundaries and cause models to focus on the most salient item rather than the full meal~\cite{furtado2020human,jagadesh2025enhancing}.

Practical dietary assessment requires more than coarse food classification. Fine-grained attributes such as subcategory and cooking style can affect nutritional interpretation; for example, distinguishing poultry from beef or grilled from roasted preparation changes the meaning of a meal log. Existing discriminative models, often built on CNN or Transformer backbones, typically treat these attributes as independent outputs and provide no guarantee of hierarchical consistency. Vision-language models (VLMs) offer stronger open-ended reasoning, but their generated outputs may include synonyms, overly specific dish names, or labels outside the dataset taxonomy, complicating evaluation and deployment~\cite{romero2025vision,lin2025vl}. These challenges highlight the need for predictions that are structured, ontology-compliant, and consistent across hierarchy levels.

\begin{figure}[t]
  \centering
  \includegraphics[width=\linewidth]{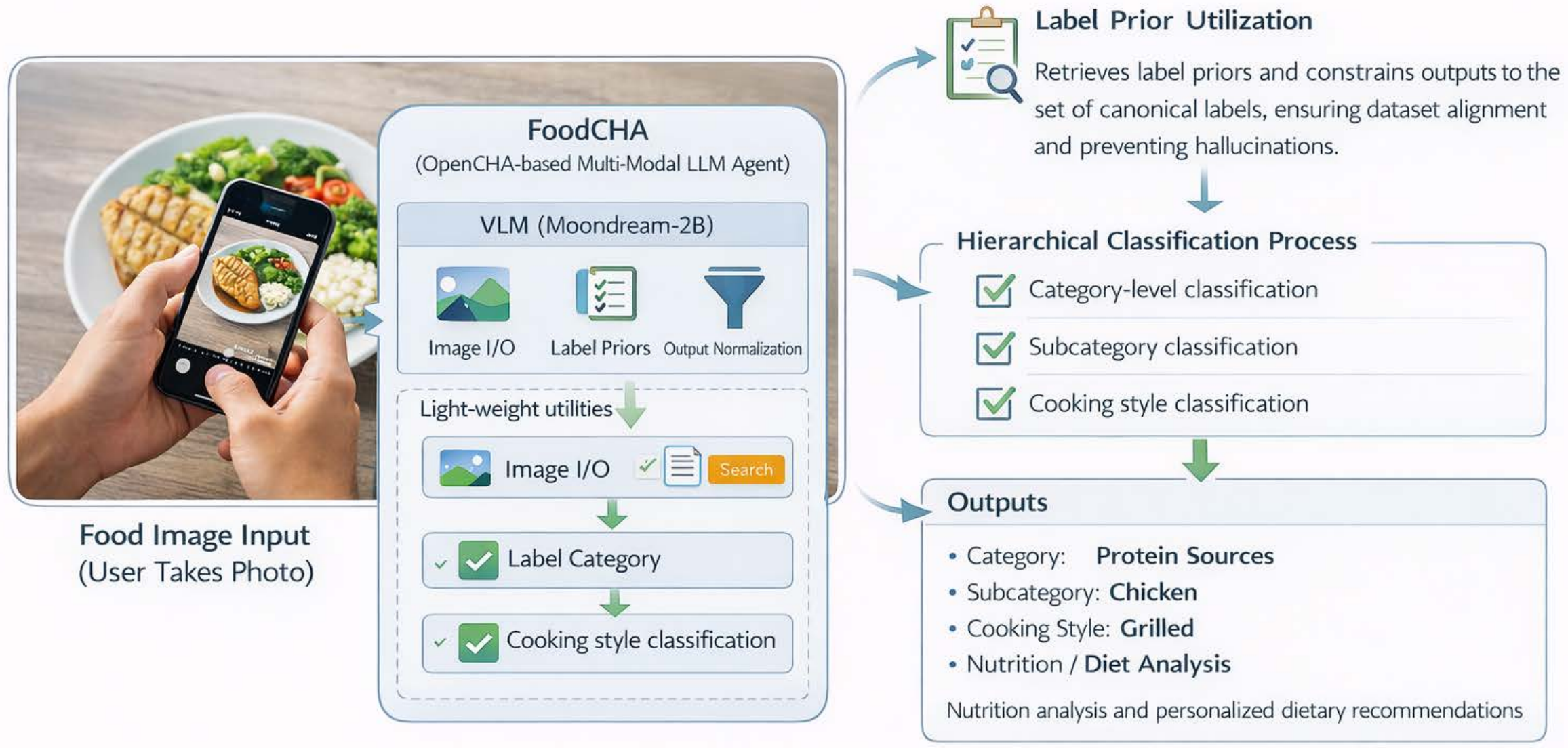}
  \caption{Overview of the FoodCHA framework.}
  \label{fig:system_overview_pic}
   \vspace{-0.2in}
\end{figure}

Addressing these limitations requires a reasoning framework that explicitly enforces hierarchical dependencies and validates intermediate outputs. Agentic orchestration provides a natural solution: by decomposing perception and reasoning into staged, tool-guided actions, predictions become structured, ontology-compliant, and hierarchy-consistent~\cite{abbasian2023conversational}. Such an approach allows fine-grained attributes, including subcategory and cooking style, to be inferred conditionally on preceding stages, reducing hallucination and enabling reliable multi-attribute dietary logging.

We propose FoodCHA (Fig.~\ref{fig:system_overview_pic}), a hierarchical VLM agent for staged prediction of food attributes. We focus on the FoodNExTDB taxonomy~\cite{romero2025vision}, which defines 10 high-level categories, 62 subcategories, and 9 cooking styles. Thus, FoodCHA targets ontology-constrained recognition within a fixed label space rather than open-world food discovery. FoodCHA predicts category, subcategory, and cooking style sequentially, restricts candidates using the dataset hierarchy, and normalizes free-form outputs to canonical labels. Compared to Food-Llama-3.2-11B-Vision-Instruct~\cite{cheng2024domainadaptivemllm,adaptllm_food_llama_hf}, FoodCHA improves category and subcategory recognition precision by 13.9\% and 38.2\%, respectively, and achieves a 153.2\% gain in cooking-style classification precision.

In summary, this work makes three key contributions: (1) formulating ontology-constrained, fine-grained food recognition for diet tracking, ensuring category, subcategory, and cooking style predictions are canonical and hierarchically consistent; (2) introducing FoodCHA, a multi-modal agentic model that decomposes recognition into a three-stage hierarchical process with dataset-constrained candidates and lightweight validation, recovery, and normalization tools; and (3) demonstrating that FoodCHA significantly outperforms state-of-the-art learning methods.

\section{Related Work}
\subsection{Deep Learning Models for Food Recognition}

Food recognition has traditionally been formulated as a discriminative visual classification task mapping an image to a fixed label set. Convolutional backbones such as ResNet are widely used for their strong inductive biases and stable optimization, while attention-enhanced variants like SENet and SGLANet improve sensitivity to informative regions and subtle ingredient cues~\cite{jagadesh2025enhancing}. More recently, transformer-based models and CNN-transformer hybrids have been explored to capture long-range dependencies. Robustness-focused designs such as NoisyViT introduce structured perturbations to mitigate overfitting under real-world noise~\cite{ghosh2025improving}. In parallel, retrieval-based pipelines offer an alternative for fine-grained recognition by matching query images against a labeled gallery and refining results via reranking. For example, CVNet replaces classical geometric verification with correlation-based verification to enhance instance-level retrieval~\cite{lee2022correlation}. Nevertheless, these approaches remain limited for nutrition-focused dietary understanding. Discriminative models often predict category, subcategory, and cooking style via separate heads without enforcing hierarchical constraints, allowing errors to propagate across levels and reducing performance on long-tailed subcategories or attributes with weak or partially occluded visual cues. Retrieval-based methods require maintaining exhaustive, up-to-date reference galleries and do not naturally produce structured, multi-attribute predictions from a single inference. These limitations highlight the need for inference that leverages semantic priors, enforces hierarchical consistency, and yields canonical labels. 

\subsection{Large Language Models for Food Recognition}
Recent work has begun leveraging large language models (LLMs) and vision-language models (VLMs) for food understanding, either by building multimodal assistants or training domain-specific food models. FoodLLM couples an LLM backbone with segmentation components to support recognition, ingredient understanding, nutrition estimation, and interactive assistance, typically relying on multi-stage training across diverse datasets~\cite{yin2024foodlmmversatilefoodassistant}. Food-Llama-3.2-11B-Vision-Instruct is a food-adapted VLM baseline, while Moondream-2B, Qwen2.5-VL-7B, and InternVL3-8B represent general VLM baselines with different model sizes and reasoning capabilities~\cite{cheng2024domainadaptivemllm,adaptllm_food_llama_hf,moondream2_hf,qwen25vl2025,zhu2025internvl3}. In addition, FoodSky improves the depth and reliability of food and nutrition knowledge in text-based question answering by training on a curated food corpus, highlighting the benefits of specialization for dietetic reasoning and guidance~\cite{ZHOU2025101234}. Advances in visual backbones, such as ResV-Mamba, further demonstrate that stronger state-space architectures can enhance fine-grained food classification on curated benchmarks~\cite{chen2024res}. Despite these advances, none of these approaches directly address the core challenge of real-world dietary logging: producing structured, dataset-constrained predictions for hierarchical attributes from images. Open-ended model outputs can drift from canonical labels, and predictions across category, subcategory, and cooking style must remain consistent. Motivated by this gap, we propose FoodCHA, a tool-augmented LLM agent that integrates semantic priors and hierarchical constraints to convert open-ended model responses into a controllable, dataset-aligned decision process.

\section{FoodCHA Design}
\label{sec:design}

\begin{figure*}[t]
  \centering
  \includegraphics[width=0.9\textwidth]{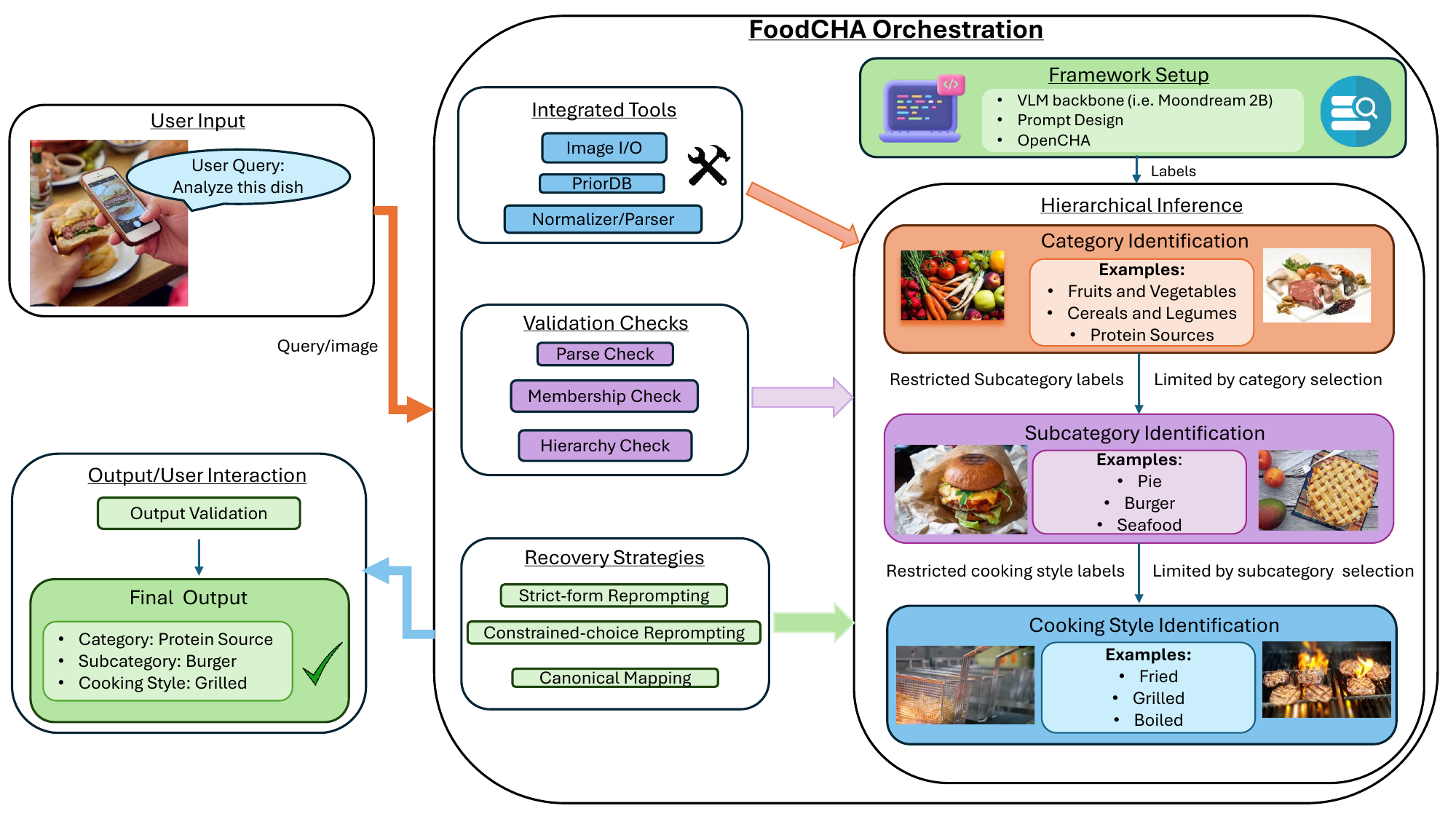}
  \caption{System-level pipeline of FoodCHA. An input image is processed by a backbone model and passed to an agent that performs hierarchical, dependency-aware decisions across category, subcategory, and cooking style. At each stage, candidate labels are constrained by the dataset taxonomy and validated, producing an ontology-compliant structured output.}
  \label{fig:system_design_pic}
\end{figure*}

\subsection{System Overview}
\label{sec:design_overview}
Figure~\ref{fig:system_design_pic} illustrates the end-to-end FoodCHA pipeline, which transforms a user-provided image into a structured dietary record. FoodCHA models recognition as a hierarchical decision sequence aligned with the dataset taxonomy. Given an input image, our agentic framework first predicts the high-level category and subsequently selects the subcategory and cooking style conditioned on prior decisions.
At each stage, candidate labels are constrained by the dataset hierarchy, ensuring that subsequent predictions remain taxonomy-valid. Integrated validation tools, including structured parsing, membership verification, and hierarchy checks, enforce canonical formatting and guarantee ontology compliance throughout the process. The final output is a deterministic $(\textit{category}, \textit{subcategory}, \textit{cooking style})$ triple that is directly displayable and suitable for structured storage.

In short, FoodCHA is guided by three primary objectives:
\begin{itemize}
  \item \textbf{Controllability:} Restrict outputs to a predefined label set and maintain a predictable, machine-parseable format.
  \item \textbf{Hierarchical Consistency:} Ensure that subcategory and cooking style predictions are valid under the dataset taxonomy and consistent with preceding decisions.
  \item \textbf{Robustness:} Reduce hallucinations, label drift, and formatting errors through constraint enforcement and recovery mechanisms.
\end{itemize}

This design enables ontology-compliant dietary logging without increasing backbone complexity. By enforcing hierarchical constraints during inference rather than through additional model capacity, FoodCHA prevents invalid label combinations while maintaining computational efficiency.

\subsection{User Input}
\label{sec:user_input}
FoodCHA takes a user-provided meal image as input and processes it under natural visual variation such as viewpoint, lighting, and background changes. Each image is mapped to a machine-parseable dietary record that can be aggregated over time for longitudinal tracking and downstream nutritional analysis.

\subsection{Framework Setup}
\label{sec:framework_setup}
Figure~\ref{fig:system_design_pic} summarizes the framework components used to process each input image, including the VLM backbone, prompting scheme, and orchestration logic.

\subsubsection{VLM Backbone}

FoodCHA uses a VLM backbone for image-text reasoning and is compatible with any model that supports structured prompting. Unlike CNN-based pipelines that perform a single closed-set prediction, FoodCHA performs staged inference in which each prediction restricts the candidate set for the next stage. In our experiments, we use Moondream-2B to balance recognition quality and computational cost. Although FoodCHA requires multiple model calls per image, each call operates over a restricted label space and is followed by deterministic validation.

\subsubsection{Prompt Design}
\label{sec:prompt_design}

FoodCHA uses stage-specific prompt templates to enforce structured, ontology-compliant outputs. At each stage, we provide the stage candidate set returned by PriorDB, require strict JSON with exactly one field, and disallow any additional text. This design minimizes free-form drift and enables deterministic parsing, membership checks, and hierarchy validation. FoodCHA performs hierarchical inference using stage-specific prompts:

\subsubsection{OpenCHA} 
FoodCHA is implemented on top of OpenCHA~\cite{abbasian2025conversational}, an agent framework designed for multi-step, multi-modal workflows with tool use. OpenCHA enables conditional branching, sequential stage orchestration, and integration of the lightweight utilities above. This ensures that each stage of inference produces ontology-valid, hierarchy-consistent, and strictly structured outputs. In combination, the VLM backbone, deterministic utilities, and OpenCHA orchestration allow FoodCHA to achieve controlled, reproducible, and expert-aligned predictions for hierarchical food recognition.

\subsection{Hierarchical Inference}
\label{sec:hier_recognition}
Forming the central prediction pathway in Figure~\ref{fig:system_design_pic}, FoodCHA only takes in the user's image as input throughout the whole process and performs hierarchical inference across three sequential stages: category, subcategory, and cooking style. Each stage predicts a single label selected from an explicitly defined candidate set derived from the dataset taxonomy. By conditioning downstream decisions on upstream predictions and restricting candidates accordingly, FoodCHA guarantees taxonomy-valid outputs and mitigates a common failure mode of free-form VLM prompting, where plausible text does not correspond to canonical labels. Within our hierarchical inference process, the user does not provide additional input between stages. As such, inference proceeds as follows:

\begin{itemize}
    \item \textbf{Stage 1 — Category.} The model predicts the high-level food category from a predefined label set. Because category cues are typically global and visually salient, this stage establishes the search space for subsequent decisions. For example, if a user provides an image of a burger dish, Stage 1 correctly identifies it as a protein source rather than a vegetable or a beverage. This initial stage is critical for narrowing the candidate set of food labels in subsequent stages. Errors at this step can propagate downstream, preventing the correct label from being selected later in the recognition process.
    
    \item \textbf{Stage 2 — Subcategory.} Conditioned on the predicted category, FoodCHA automatically retrieves valid subcategory labels from the dataset hierarchy and restricts prediction to this subset. This prevents drift into irrelevant or non-canonical labels. Continuing with the burger example, once the dish is classified "Protein Sources" in Stage~1, Stage~2 can correctly identify it as a burger. Selecting “Protein Sources” as the category restricts the subcategory candidates to valid protein-based options, such as lamb chops, steaks, fish, or chicken.
    
    \item \textbf{Stage 3 — Cooking Style.} Conditioned on both category and subcategory, the model autonomously predicts the cooking style from the corresponding valid set. Deferring this decision allows the model to focus on preparation-specific cues. Once the food item has been classified as a protein source and specifically as a burger, Stage~3 further restricts the cooking-style candidates to valid options, such as grilled, fried, or baked. This allows the system to correctly identify the dish as a grilled burger.
\end{itemize}

\begin{tcolorbox}[]
\footnotesize

\textbf{Stage 1: Category Selection}

Task: Choose the single best category label for the food in the image.\\
Valid labels: \{candidates\_txt\}\\
Output (strict JSON): \texttt{\{"category": "<LABEL>"\}}\\
Constraint: Output must exactly match one candidate label (or \texttt{\{"unknown"\}}).\\[0.4em]

\textbf{Stage 2: Subcategory Selection}

Context: category = "\{category\}"\\
Task: Choose the best subcategory under the given category.\\
Valid labels: \{candidates\_txt\}\\
Output (strict JSON): \texttt{\{"subcategory": "<LABEL>"\}}\\
Constraint: Exact match to a candidate label.\\[0.4em]

\textbf{Stage 3: Cooking Style Selection}

Context: category = "\{category\}", subcategory = "\{subcategory\}"\\
Task: Choose the best cooking style label.\\
Valid labels: \{candidates\_txt\}\\
Output (strict JSON): \texttt{\{"cooking\_style": "<LABEL>"\}}\\
Constraint: Exact match to a candidate label.

\end{tcolorbox}

At each stage, only labels that are valid descendants in the taxonomy are considered, preventing semantically invalid combinations by design. Constraining the label space in successive stages reduces hallucinations and increases the likelihood that the resulting category, subcategory, and cooking-style combination is logically consistent. Without hierarchical inference, label combinations could be both incorrect and internally contradictory, reducing user trust in the system.

\subsection{Integrated Tools} 
\label{sec:integrated_tools}
FoodCHA incorporates three deterministic utilities to support controlled inference:

\begin{itemize}
    \item \textbf{PriorDB:} Encodes the FoodNExTDB ontology as a typed hierarchy (e.g., \texttt{category} $\rightarrow$ \texttt{subcategory} $\rightarrow$ \texttt{cooking\_style}) and returns stage-specific candidate sets $\mathcal{C}_s$. Given upstream decisions, it provides only valid next-stage labels and optional aliases, reducing open-vocabulary drift and invalid label combinations.
    
    \item \textbf{Normalizer/Parser:} Extracts canonical labels from model outputs, enforces fixed keys, normalizes surface variations, and maps responses to valid labels through exact matches, synonym lookup, or unique approximate matches above threshold $\tau$.
    
    \item \textbf{ImageIO:} Standardizes image ingestion, including resizing, cropping, color-space conversion, and optional compression, ensuring reproducible inputs across repeated model calls.
\end{itemize}

\subsection{Validation Checks}
\label{sec:validation}
FoodCHA validates each stage before proceeding. Let $\mathcal{O}$ denote the ontology of allowed labels, and let $\pi(\cdot)$ be a parser that extracts a candidate label or item-level label set from the model response. Each stage enforces three checks:

\begin{itemize}
    \item \textbf{Parse check:} Ensures exactly one label per field for each predicted food item (e.g., \texttt{category}, \texttt{subcategory}, \texttt{cooking\_style}). For mixed meals, the parser accepts a bounded list of item-level records, where each record follows the same schema.
   \item \textbf{Membership check:} The extracted label $\hat{y}$ must belong to the stage-specific candidate set $\mathcal{C} \subseteq \mathcal{O}$. Formally, accept if $\hat{y} \in \mathcal{C}$, otherwise reject. $\mathcal{C}$ is conditioned on the current stage (e.g., top-level categories vs. subcategories given the category).
   \item \textbf{Hierarchy check:} Ensures parent--child consistency using a parent map $g: \mathcal{O} \rightarrow 2^{\mathcal{O}}$, where $g(y)$ returns the set of valid children of $y$. If the upstream label is $y^{(k)}$, the downstream prediction $y^{(k+1)}$ is accepted only if $y^{(k+1)} \in g(y^{(k)})$.
\end{itemize}

\subsection{Recovery strategies} 
As shown with the recovery branch illustrated in Figure~\ref{fig:system_design_pic}, FoodCHA also has bounded recovery after each stage to further ensure robustness and proper outputs. When a check fails, FoodCHA applies bounded recovery with a maximum of $R$ retries per stage. Each retry increases the strictness of output control:

\begin{enumerate}
    \item \textbf{Strict-format re-prompting:} Enforces an explicit output schema, blocking explanations or extra text to correct parse failures.
    \item \textbf{Constrained-choice re-prompting:} Restricts the model to the stage candidate set $\mathcal{C}$ and requires an exact single-label selection. Can be combined with constrained decoding to limit next-token generation to valid label tokens.
    \item \textbf{Canonical mapping:} Applies deterministic normalization for near-miss outputs (e.g., spelling, pluralization, hyphenation) using lowercasing, punctuation stripping, synonym mapping, and string similarity threshold $\tau$. Mapping is applied only if the nearest valid label is unique and within the threshold.
\end{enumerate}

If validation still fails after $R$ retries, FoodCHA outputs a structured \texttt{unknown} label with an error code indicating the failed check and the recovery mode attempted. This explicit reporting prevents silent corruption, enables logging, and supports debugging while maintaining predictable latency.
\newline

\subsection{Output/User Interaction}
\label{sec:user_interaction}

Corresponding to the terminal output stage in Figure~\ref{fig:system_design_pic}, FoodCHA produces structured, taxonomy-compliant predictions that can be displayed to users and appended to meal logs. For single-item meals, FoodCHA returns one label triple. For mixed meals, it can return a bounded list of item-level triples so that secondary food items are not forced into a single dominant prediction. This format supports aggregation, longitudinal trend analysis, and downstream nutrition computation more reliably than free-form text.

To illustrate the staged process, consider an input meal photo containing a burger. FoodCHA performs three sequential calls, each time restricting the candidate set to valid descendants in the taxonomy. Stage~1 selects the coarse category from the global set, Stage~2 selects a subcategory from only those valid under the chosen category, and Stage~3 selects a cooking style valid for the selected (category, subcategory) pair:
\begin{tcolorbox}[]
\footnotesize

\textbf{Input:} Burger meal photo

\textbf{Stage 1 (Category)}\\
Candidates: \{Protein Sources, Grains, Vegetables, \dots\}\\
Output: \texttt{\{"category": "Protein Sources"\}}

\textbf{Stage 2 (Subcategory)}\\
Candidates (Protein Sources): \{Burger, Steak, Fish, Poultry, \dots\}\\
Output: \texttt{\{"subcategory": "Burger"\}}

\textbf{Stage 3 (Cooking Style)}\\
Candidates (Protein Sources, Burger): \{Grilled, Fried, Oven Baked, none, \dots\}\\
Output: \texttt{\{"cooking\_style": "Grilled"\}}

\textbf{Final Prediction}\\
\texttt{\{"category": "Protein Sources", "subcategory": "Burger", "cooking\_style": "Grilled"\}}

\end{tcolorbox}

By constraining outputs to canonical labels, FoodCHA reduces label variance and enforces a consistent schema for downstream processing and comparison across meals and users.

\section{Experimental Analysis}
\label{sec:results}
\subsection{Experimental Setup}

\textbf{Dataset.} We evaluate FoodCHA on FoodNExTDB~\cite{romero2025vision}, a hierarchical food-recognition benchmark with three prediction targets: \textit{Category} (Stage~1), \textit{Subcategory} (Stage~2), and \textit{Cooking Style} (Stage~3). It consists of 10 high-level categories, 62 subcategories, and 9 cooking styles. We split the dataset into 6,483 training images, 1,389 validation images, and 1,390 test images. FoodNExTDB is well suited for evaluating FoodCHA because its hierarchical structure aligns with staged outputs, its multi-annotator design captures the inherent ambiguity of consumer meal photos, and its scale supports comprehensive pipeline analysis. Multiple expert annotations per item enable expert-aligned evaluation under inter-annotator disagreement.

\textbf{Task.} The goal is to predict canonical labels from the FoodNExTDB taxonomy. FoodCHA performs staged inference: it first predicts the Category, restricts Subcategory candidates to those valid under the chosen Category, and finally predicts Cooking Style conditioned on previous stages. This hierarchical design enforces structural validity and ensures reliable logging for downstream dietary tracking and nutrition analysis.

\textbf{State-of-the-art Baselines.} We compare FoodCHA against both CNN- and VLM-based methods. CNN baselines include SENet~\cite{hu2018squeeze}, SGLANet~\cite{abiyev2024automatic}, and CVNet~\cite{lee2022correlation}. VLM baselines include Food-Llama-3.2-11B-Vision-Instruct~\cite{cheng2024domainadaptivemllm}, InternVL3-8B~\cite{zhu2025internvl3}, Moondream-2B~\cite{moondream2_hf}, and Qwen2.5-VL-7B~\cite{qwen25vl2025}. All VLM baselines are evaluated under the same FoodNExTDB label space and image preprocessing. FoodCHA uses batch size 1, strict JSON outputs, and a retry budget of 3 per stage. The same canonical parsing and normalization rules are applied during evaluation. One-shot VLM baselines are prompted to predict category, subcategory, and cooking style in a structured format, but they do not receive stage-wise candidate restriction, hierarchy validation, or bounded recovery.

\textbf{Evaluation Metrics.} We report Precision, Recall, and F1-score for each stage. To account for inter-annotator ambiguity, we also report Expert-Weighted Recall (EWR)~\cite{romero2025vision}:

$$
\mathrm{EWR}_i
=
\frac{
\sum_{k=1}^{L_i} W_{\mathrm{VLM}}\!\left(i, l_k^i\right)
}{
\sum_{k=1}^{L_i} W_{\mathrm{Nutritionists}}\!\left(i, l_k^i\right)
},
$$
where 
$L_i$ is the number of labels annotated for image $i$, 
$l_k^i$ is the $k$-th annotated label, 
$W_{\mathrm{Nutritionists}}(i,l_k^i)$ is the expert-consensus weight for label $l_k^i$, and 
$W_{\mathrm{VLM}}(i,l_k^i)$ is the weight assigned to the model's prediction. Since multiple annotators may disagree, EWR weights each test example according to expert consensus: predictions that match highly agreed-upon labels contribute more, while predictions on ambiguous labels contribute less. This makes EWR particularly suitable for evaluating subcategory and cooking-style predictions.

\textbf{Hardware.} All experiments were conducted on a workstation equipped with an NVIDIA RTX 4090 GPU. 
\begin{table*}[!t]
\centering
\caption{
\textbf{Overall prediction performance on FoodNExTDB.} Precision (Prec.), Recall (Rec.), F1-Score (F1.), and Expert-Weighted Recall (EWR) are reported for the three hierarchical stages: Category, Subcategory, and Cooking Style. 
}
\label{tab:overall_results}
\setlength{\tabcolsep}{3.0pt}
\renewcommand{\arraystretch}{1.05}
\scriptsize
\begin{tabular}{l l
c c c c
| c c c c
| c c c c}
\hline
\multirow{2}{*}{\textbf{Family}} & \multirow{2}{*}{\textbf{Method}}
& \multicolumn{4}{c|}{\textbf{Category (Stage 1)}}
& \multicolumn{4}{c|}{\textbf{Subcategory (Stage 2)}}
& \multicolumn{4}{c}{\textbf{Cooking Style (Stage 3)}} \\
& & \textbf{Prec.} & \textbf{Rec.} & \textbf{F1.} & \textbf{EWR}
  & \textbf{Prec.} & \textbf{Rec.} & \textbf{F1.} & \textbf{EWR}
  & \textbf{Prec.} & \textbf{Rec.} & \textbf{F1.} & \textbf{EWR} \\
\hline\hline

CNN baselines
& SENet
& 0.760 & 0.648 & 0.700 & 0.605
& 0.590 & 0.511 & 0.548 & 0.410
& 0.613 & \textbf{0.588} & 0.600 & 0.560 \\
& SGLANet
& 0.740 & 0.620 & 0.675 & 0.622
& 0.559 & 0.484 & 0.519 & 0.451
& 0.610 & 0.572 & 0.590 & 0.588 \\
& CVNet
& 0.280 & 0.270 & 0.275 & 0.255
& 0.060 & 0.070 & 0.065 & 0.060
& 0.310 & 0.330 & 0.320 & 0.290 \\
\hline

VLM baselines
& Moondream-2B
& 0.781 & 0.759 & 0.770 & 0.710
& 0.652 & 0.616 & 0.633 & 0.602
& 0.267 & 0.226 & 0.245 & 0.203 \\
& Food-Llama-3.2-11B
& 0.792 & 0.742 & 0.766 & 0.742
& 0.571 & 0.468 & 0.514 & 0.446
& 0.344 & 0.318 & 0.330 & 0.286 \\
& InternVL3-8B
& 0.748 & 0.701 & 0.724 & 0.688
& 0.506 & 0.402 & 0.448 & 0.361
& 0.301 & 0.287 & 0.294 & 0.241 \\
& Qwen2.5-VL-7B
& 0.554 & 0.445 & 0.494 & 0.435
& 0.159 & 0.077 & 0.104 & 0.032
& 0.213 & 0.230 & 0.221 & 0.183 \\
\hline

VLM (ours)
& \textbf{FoodCHA}
& \textbf{0.902} & \textbf{0.761} & \textbf{0.826} & \textbf{0.820}
& \textbf{0.789} & \textbf{0.579} & \textbf{0.668} & \textbf{0.648}
& \textbf{0.871} & 0.556 & \textbf{0.679} & \textbf{0.640} \\
\hline
\end{tabular}
\end{table*}

\subsection{Performance Across Stages}

Table~\ref{tab:overall_results} compares FoodCHA with CNN and VLM baselines across all three stages. FoodCHA achieves the strongest overall performance, with gains persisting from coarse category recognition to fine-grained cooking-style identification.

\textbf{Category (Stage~1).} FoodCHA attains 0.820 EWR and 0.826 F1, with precision reaching 0.902. Since category cues are often global and visually salient, the strongest baselines remain relatively competitive at this level. However, stronger category prediction provides a more reliable starting point for downstream hierarchy-constrained decisions.

\textbf{Subcategory (Stage~2).} FoodCHA reaches 0.648 EWR and 0.668 F1. This stage is more difficult because visually similar dishes must be separated within a denser label space. FoodCHA improves performance by restricting prediction to category-valid subcategories, removing irrelevant alternatives from consideration.

\textbf{Cooking Style (Stage~3).} FoodCHA shows its clearest advantage at the cooking-style stage, reaching 0.640 EWR and 0.679 F1 with 0.871 precision. Cooking style depends on subtle preparation cues and consistency with earlier decisions, making one-shot VLMs vulnerable to drift. FoodCHA narrows the output space before predicting style, reducing hallucination and invalid label combinations. The fact that larger one-shot VLMs still underperform suggests that backbone scale alone is insufficient for reliable taxonomy-valid food recognition.
\begin{figure}[]
  \centering
  \includegraphics[width=.9\linewidth]{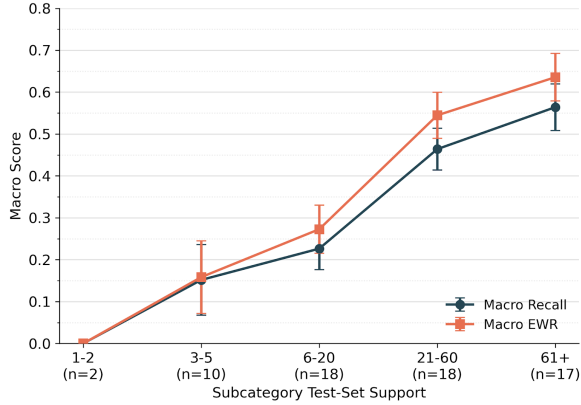}
  \caption{\textbf{Subcategory analysis.} Macro Recall and Macro EWR over subcategory bins defined by test-set support.}
  \label{fig:subcat_longtail}
\end{figure}

\subsection{Fine-Grained Dish Element Analysis}

\textbf{Subcategory Performance.} 
Figure~\ref{fig:subcat_longtail} illustrates FoodCHA's performance on subcategory predictions as a function of label frequency. The x-axis groups subcategories by their occurrence in the dataset, and the y-axis shows Macro Recall and Macro EWR with 95\% bootstrap confidence intervals. Performance decreases for rare subcategories due to limited examples and visual similarity with other classes. Notably, the gap between Recall and EWR increases for these tail classes, indicating that many errors are near misses rather than completely incorrect predictions. For example, within the Fast Food category, visually similar items such as Burgers and Sandwiches can be confused under occlusion, similar wraps, or mixed fillings. 

Predictions that are marked as incorrect by Recall may still receive partial credit under EWR because expert annotators also find these labels ambiguous. This behavior demonstrates the benefit of FoodCHA's staged, candidate-restricted inference, which constrains predictions to valid subcategories in the ontology. As a result, errors tend to remain close to expert consensus. Nevertheless, subcategory prediction remains the most challenging stage, and early mistakes can propagate to downstream tasks, such as cooking style classification, limiting recovery even when fine-grained visual cues are available.

\begin{figure}[]
  \centering
  \includegraphics[width=.9\linewidth]{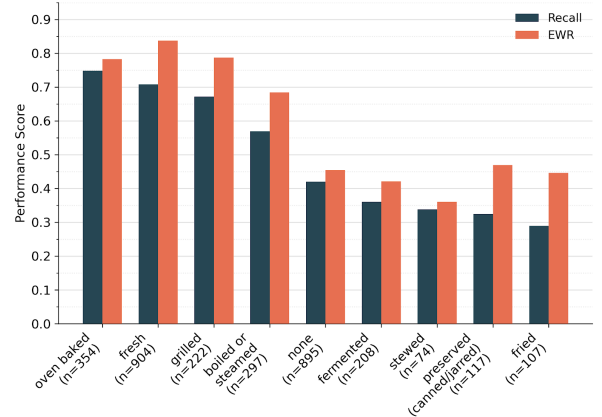}
  \vspace{-0.3cm}
  \caption{\textbf{Cooking-style analysis.} Recall and EWR for each cooking-style label.}
  \label{fig:style_breakdown}
\end{figure}

\textbf{Cooking Style Performance.}
Figure~\ref{fig:style_breakdown} presents a per-class analysis of cooking style predictions. Cooking style recognition shows the largest relative improvements under FoodCHA, as unconstrained VLMs often drift when preparation cues are subtle or overlapping. Among all styles, \textit{Oven Baked} achieves the strongest performance, likely because this preparation method produces clear and repeatable visual signatures such as uniform browning, crisp edges, and dry surface texture. \textit{Grilled} also performs strongly due to distinctive grill marks and localized charring. Similarly, \textit{Fresh} ranks highly because raw or minimally processed ingredients preserve characteristic textures and colors, and the absence of browning or crust formation provides an informative visual signal. \textit{Boiled Or Steamed} also shows relatively strong performance, as its moisture-rich appearance and low browning surfaces visually distinguish it from high heat cooking styles. 

Performance declines for the remaining categories. The \textit{None} class is particularly challenging because it represents a heterogeneous group of food items that do not clearly belong to any defined cooking style, which increases the likelihood of confusion with other categories. For styles such as \textit{Fermented}, \textit{Stewed}, and \textit{Preserved}, visual cues are often subtle and may overlap with other preparation styles, especially when dishes are combined with sauces. Although \textit{Fried} foods may appear to have strong visual signals, their characteristic texture cues such as crispness or bubbling surfaces are not always visible in natural images. However, it is worth noting that both \textit{Preserved} and \textit{Fried} exhibit substantially higher EWR than Recall, indicating that disagreements frequently arise among semantically related styles rather than arbitrary labels. This pattern further suggests that FoodCHA tends to produce cautious, ontology-aligned predictions that remain close to expert consensus.

\subsection{Ablation Study}

Table~\ref{tab:ablation} isolates the effects of the backbone, hierarchical prompting, and OpenCHA validation. The Moondream-2B baseline uses the same backbone but predicts without staged hierarchy constraints. FoodCHA w/o OpenCHA keeps hierarchical prompting and candidate restriction, but removes explicit parsing, validation, and bounded recovery. Full FoodCHA combines hierarchy-aware candidate restriction with OpenCHA-based validation.

FoodCHA improves over Moondream-2B across all stages, with the largest gain at cooking style, where EWR rises from 0.203 to 0.640. This reflects the benefit of progressively restricting candidate sets to ontology-valid descendants. Comparing FoodCHA w/o OpenCHA to the full system shows that validation matters most at the finest-grained level: cooking-style EWR increases from 0.529 to 0.640. OpenCHA reduces formatting errors, invalid labels, and hierarchy drift through parsing, membership checks, hierarchy checks, and bounded recovery. These results show that both hierarchy-aware prompting and validation are needed for reliable fine-grained prediction.

\begin{table}[!t]
\centering
\caption{Ablation study across prediction stages. EWR is reported for each stage.}
\label{tab:ablation}
\scriptsize
\setlength{\tabcolsep}{4pt}
\renewcommand{\arraystretch}{1.10}
\resizebox{\columnwidth}{!}{%
\begin{tabular}{lccc}
\hline
Method & Category EWR & Subcategory EWR & Cooking Style EWR \\
\hline
\shortstack[l]{Moondream-2B} & 0.710 & 0.602 & 0.203 \\
\shortstack[l]{FoodCHA w/o OpenCHA} & 0.764 & 0.631 & 0.529 \\
\shortstack[l]{FoodCHA} & \textbf{0.820} & \textbf{0.648} & \textbf{0.640} \\
\hline
\end{tabular}%
}
\end{table}

\subsection{Overhead Analysis}

Table~\ref{tab:latency} reports end-to-end latency per image (seconds) and summarizes the distribution using the median (p50) and tail percentiles (p90, p95). Here, p50 denotes the median latency, while p90 and p95 correspond to the latencies below which 90\% and 95\% of samples complete, respectively.

Overall, FoodCHA increases median latency from 1.29\,s for one-shot Moondream-2B to 3.19\,s due to two additional conditioned calls introduced by hierarchical staging. Despite this overhead, FoodCHA remains substantially faster than larger one-shot VLMs, including Qwen2.5-VL-7B (9.97\,s) and Food-Llama-3.2-11B (15.65\,s). This result suggests that structured, hierarchy-aware inference can achieve reliable canonical label selection more efficiently than scaling backbone capacity alone. The additional computation is therefore devoted to enforcing taxonomy validity and improving expert-aligned agreement through candidate restriction.

\begin{table}[t]
\centering
\caption{End-to-end inference latency. Latency is reported in seconds per image.}
\label{tab:latency}
\scriptsize
\setlength{\tabcolsep}{4.0pt}
\renewcommand{\arraystretch}{1.05}
\begin{tabular}{l c c c c}
\hline
\textbf{Model} & \textbf{Avg} & \textbf{p50} & \textbf{p90} & \textbf{p95} \\
\hline
Moondream 2B & 1.50 & 1.29 & 2.09 & 2.61 \\
FoodCHA & 3.57 & 3.19 & 4.11 & 4.58 \\
Qwen2.5-VL-7B & 9.43 & 9.97 & 11.39 & 12.03 \\
Food-Llama-3.2-11B & 14.80 & 15.65 & 17.88 & 18.90 \\
\hline
\end{tabular}
\end{table}

\section{Conclusion}
We introduce FoodCHA, an OpenCHA-based agent pipeline that reformulates food recognition as a sequence of hierarchy-constrained decisions across category, subcategory, and cooking style. By leveraging the dataset taxonomy to restrict candidate sets at each stage, FoodCHA produces canonical, hierarchy-consistent labels and mitigates drift inherent in open-ended VLM outputs. Evaluated on FoodNExTDB~\cite{romero2025vision}, FoodCHA consistently outperforms state-of-the-art CNN and VLM baselines, including SENet, SGLANet, Moondream-2B, Qwen2.5-VL-7B, and Food-Llama-3.2-11B-Vision-Instruct, across all three hierarchy levels. Its largest gains appear at the cooking-style stage, where staged conditioning substantially improves both F1 and EWR. Importantly, these gains are realized without increasing backbone complexity, demonstrating that structured, taxonomy-aware inference can deliver reliable, expert-aligned diet logging across diverse platforms.

\nocite{*}
\bibliographystyle{plain} 
\bibliography{references}    

\end{document}